\documentclass{article} 
\usepackage{nips12submit_e,times}
\usepackage{epsf}
\usepackage{subfigure}
\usepackage{tabularx}
\usepackage{amssymb}
\usepackage{subfigure}
\usepackage{graphicx}
\usepackage{wrapfig}
\usepackage{amsmath}
\usepackage{url}
\usepackage{algorithm}
\usepackage{algorithmic}
\usepackage{enumitem}
\usepackage{epstopdf}
\usepackage[T1]{fontenc}
\usepackage{eqparbox}

\title{Training Neural Networks with Stochastic Hessian-Free Optimization}

\author{
Ryan Kiros \\ 
Department of Computing Science\\
University of Alberta\\
Edmonton, AB, Canada \\
\texttt{rkiros@ualberta.ca} \\
}

\nipsfinalcopy 

\begin{document}

\maketitle

\begin{abstract}

Hessian-free (HF) optimization has been successfully used for training deep autoencoders and recurrent networks. HF uses the conjugate gradient algorithm to construct update directions through curvature-vector products that can be computed on the same order of time as gradients. In this paper we exploit this property and study stochastic HF with gradient and curvature mini-batches independent of the dataset size. We modify Martens' HF for these settings and integrate dropout, a method for preventing co-adaptation of feature detectors, to guard against overfitting. Stochastic Hessian-free optimization gives an intermediary between SGD and HF that achieves competitive performance on both classification and deep autoencoder experiments.

\end{abstract}

\section{Introduction}

Stochastic gradient descent (SGD) has become the most popular algorithm for training neural networks. Not only is SGD simple to implement but its noisy updates often leads to solutions that are well-adapt to generalization on held-out data \cite{bottou2011}. Furthermore, SGD operates on small mini-batches potentially allowing for scalable training on large datasets. For training deep networks, SGD can be used for fine-tuning after layerwise pre-training \cite{bengio2007greedy} which overcomes many of the difficulties of training deep networks. Additionally, SGD can be augmented with dropout \cite{hinton2012improving} as a means of preventing overfitting.

There has been recent interest in second-order methods for training deep networks, partially due to the successful adaptation of Hessian-free (HF) by \cite{martens2010deep}, an instance of the more general family of truncated Newton methods. Second-order methods operate in batch settings with less but more substantial weight updates. Furthermore, computing gradients and curvature information on large batches can easily be distributed across several machines. Martens' HF was able to successfully train deep autoencoders without the use of pre-training and was later used for solving several pathological tasks in recurrent networks \cite{martens2011learning}.

HF iteratively proposes update directions using the conjugate gradient algorithm, requiring only curvature-vector products and not an explicit computation of the curvature matrix. Curvature-vector products can be computed on the same order of time as it takes to compute gradients with an additional forward and backward pass through the function's computational graph \cite{pearlmutter1994fast, schraudolph2002fast}. In this paper we exploit this property and introduce stochastic Hessian-free optimization (SHF), a variation of HF that operates on gradient and curvature mini-batches independent of the dataset size. Our goal in developing SHF is to combine the generalization advantages of SGD with second-order information from HF. SHF can adapt its behaviour through the choice of batch size and number of conjugate gradient iterations, for which its behaviour either becomes more characteristic of SGD or HF. Additionally we integrate dropout, as a means of preventing co-adaptation of feature detectors. We perform experimental evaluation on both classification and deep autoencoder tasks. For classification, dropout SHF is competitive with dropout SGD on all tasks considered while for autoencoders SHF performs comparably to HF and momentum-based methods. Moreover, no tuning of learning rates needs to be done.

\section{Related work}

Much research has been investigated into developing adaptive learning rates or incorporating second-order information into SGD. \cite{lecun1998efficient} proposed augmenting SGD with a diagonal approximation of the Hessian while Adagrad \cite{duchi2010adaptive} uses a global learning rate while dividing by the norm of previous gradients in its update. SGD with Adagrad was shown to be beneficial in training deep distributed networks for speech and object recognition \cite{dean2012large}. To completely avoid tuning learning rates, \cite{schaul2012no} considered computing rates as to minimize estimates of the expectation of the loss at any one time. \cite{bordes2009sgd} proposed SGD-QN for incorporating a quasi-Newton approximation to the Hessian into SGD and used this to win one of the 2008 PASCAL large scale learning challenge tracks.
Recently, \cite{pascanu2013natural} provided a relationship between HF, Krylov subspace descent and natural gradient due to their use of the Gauss-Newton curvature matrix. Furthermore, \cite{pascanu2013natural} argue that natural gradient is robust to overfitting as well as the order of the training samples. Other methods incorporating the natural gradient such as TONGA \cite{le2007topmoumoute} have also showed promise on speeding up neural network training.

Analyzing the difficulty of training deep networks was done by \cite{glorot2010understanding}, proposing a weight initialization that demonstrates faster convergence. More recently, \cite{dauphin2013big} argue that large neural networks waste capacity in the sense that adding additional units fail to reduce underfitting on large datasets. The authors hypothesize the SGD is the culprit and suggest exploration with stochastic natural gradient or stochastic second-order methods. Such results further motivate our development of SHF. \cite{sutskever2013training} show that with careful attention to the parameter initialization and momentum schedule, first-order methods can be competitive with HF for training deep autoencoders and recurrent networks. We compare against these methods in our autoencoder evaluation.

Related to our work is that of \cite{byrd2012sample}, who proposes a dynamic adjustment of gradient and curvature mini-batches for HF with convex losses based on variance estimations. Unlike our work, the batch sizes used are dynamic with a fixed ratio and are initialized as a function of the dataset size. Other work on using second-order methods for neural networks include \cite{chapelle2011improved} who proposed using the Jacobi pre-conditioner for HF, \cite{sutskever2011generating} using HF to generate text in recurrent networks and \cite{vinyals2011krylov} who explored training with Krylov subspace descent (KSD). Unlike HF, KSD could be used with Hessian-vector products but requires additional memory to store a basis for the Krylov subspace. L-BFGS has also been successfully used in fine-tuning pre-trained deep autoencoders, convolutional networks \cite{le2011optimization} and training deep distributed networks \cite{dean2012large}. Other developments and detailed discussion of gradient-based methods for neural networks is described in \cite{bengio2012practical}.

\section{Hessian-free optimization}

In this section we review Hessian-free optimization, largely following the implementation of Martens \cite{martens2010deep}. We refer the reader to \cite{martens2012training} for detailed development and tips for using HF.

We consider unconstrained minimization of a function $f : \mathbb{R}^n \rightarrow \mathbb{R}$ with respect to parameters $\theta$. More specifically, we assume $f$ can be written as a composition $f(\theta) = L(F(\theta))$ where $L$ is a convex loss function and $F(\theta)$ is the output of a neural network with $\ell$ non-input layers. We will mostly focus on the case when $f$ is non-convex. Typically $L$ is chosen to be a matching loss to a corresponding transfer function $p(z) = p(F(\theta))$. For a single input, the $(i+1)$-th layer of the network is expressed as
\begin{eqnarray}
y_{i+1} = s_i (W_i \text{ } y_i + b_i)
\end{eqnarray}
where $s_i$ is a transfer function, $W_i$ is the weights connecting layers $i$ and $i+1$ and $b_i$ is a bias vector. Common transfer functions include the sigmoid $s_i(x) = (1 + \text{exp}(-x))^{-1}$, the hyperbolic tangent $s_i(x) = \text{tanh}(x)$ and rectified linear units $s_i(x) = \text{max}(x, 0)$. In the case of classification tasks, the loss function used is the generalized cross entropy and softmax transfer
\begin{eqnarray}
L(p(z), t) = -\sum_{j=1}^k t_j \text{ log}(p(z_j)), \hspace{3mm} p(z_j) = \text{exp}(z_j) / \sum_{l=1}^k \text{exp}(z_l)
\end{eqnarray}
where $k$ is the number of classes, $t$ is a target vector and $z_j$ the $j$-th component of output vector $z$. Consider a local quadratic approximation $M_{\theta}(\delta)$ of $f$ around $\theta$:
\begin{eqnarray}
\label{qaf}
f(\theta + \delta) \approx M_{\theta}(\delta) = f(\theta) + \nabla f(\theta)^T \delta + \frac{1}{2} \delta^T B \delta
\end{eqnarray}
where $\nabla f(\theta)$ is the gradient of $f$ and $B$ is the Hessian or an approximation to the Hessian. If $f$ was convex, then $B \succeq 0$ and equation $\ref{qaf}$ exhibits a minimum $\delta^*$. In Newton's method, $\theta_{k+1}$, the parameters at iteration $k+1$, are updated as $\theta_{k+1} = \theta_{k} + \alpha_k \delta^*_k$ where $\alpha_k \in [0,1]$ is the rate and $\delta^*_k$ is computed as
\begin{eqnarray}
\delta^*_k = - B^{-1} \nabla f(\theta_{k-1})
\end{eqnarray}
for which calculation requires $O(n^3)$ time and thus often prohibitive. Hessian-free optimization alleviates this by using the conjugate gradient (CG) algorithm to compute an approximate minimizer $\delta_k$. Specifically, CG minimizes the quadratic objective $q(\delta)$ given by
\begin{eqnarray}
q(\delta) = \frac{1}{2} \delta^T B \delta + \nabla f(\theta_{k-1})^T \delta
\end{eqnarray}
for which the corresponding minimizer of $q(\delta)$ is $-B^{-1} \nabla f(\theta_{k-1})$. The motivation for using CG is as follows: while computing $B$ is expensive, compute the product $Bv$ for some vector $v$ can be computed on the same order of time as it takes to compute $\nabla f(\theta_{k-1})$ using the R-operator \cite{pearlmutter1994fast}. Thus CG can efficiently compute an iterative solution to the linear system $B \delta_k = -\nabla (f(\theta_{k-1}))$ corresponding to a new update direction $\delta_k$. 

When $f$ is non-convex, the Hessian may not be positive semi-definite and thus equation $\ref{qaf}$ no longer has a well defined minimum. Following Martens, we instead use the generalized Gauss-newton matrix defined as $B = J^T L^{''} J$ where $J$ is the Jacobian of $f$ and $L^{''}$ is the Hessian of $L$ \footnote{While an abuse of definition, we still refer to ``curvature-vector products'' and ``curvature batches'' even when $B$ is used.}. So long as $f(\theta) = L(F(\theta))$ for convex $L$ then $B \succeq 0$. Given a vector $v$, the product $Bv =  J^T L^{''} Jv$ is computed successively by first computing $Jv$, then $L^{''} (Jv)$ and finally $J^T (L^{''} Jv)$ \cite{schraudolph2002fast}. To compute $Jv$, we utilize the R-operator. The R-operator of $F(\theta)$ with respect to $v$ is defined as
\begin{eqnarray}
\mathcal{R}_v\{F(\theta)\} = \underset{\epsilon \rightarrow 0}{\text{lim }} \frac{F(\theta + \epsilon v) - F(\theta)}{\epsilon} = Jv
\end{eqnarray}
Computing $\mathcal{R}_v\{F(\theta)\}$ in a neural network is easily done using a forward pass by computing $\mathcal{R}_v \{y_{i} \}$ for each layer output $y_i$. More specifically,
\begin{eqnarray}
\mathcal{R}_v \{y_{i+1} \} = \mathcal{R}_v \{ W_i \text{ } y_i + b_i \} s_i'= (v(W_i) y_i \text{ } + v(b_i) + W_i \mathcal{R} \{y_i \} ) s_i'
\end{eqnarray}
where $v(W_i)$ is the components of $v$ corresponding to parameters between layers $i$ and $i+1$ and $\mathcal{R} \{y_1 \} = 0$ (where $y_1$ is the input data). In order to compute $J^T (L^{''} Jv)$, we simply apply backpropagation but using the vector $L^{''} Jv$ instead of $\nabla L$ as is usually done to compute $\nabla f$. Thus, $Bv$ may be computed through a forward and backward pass in the same sense that $L$ and $\nabla f = J^T \nabla L$ are. 

As opposed to minimizing equation $\ref{qaf}$, Martens instead uses an additional damping parameter $\lambda$ with damped quadratic approximation
\begin{eqnarray}
\hat{M}_{\theta}(\delta) = f(\theta) + \nabla f(\theta)^T \delta + \frac{1}{2} \delta^T \hat{B} \delta = f(\theta) + \nabla f(\theta)^T \delta + \frac{1}{2} \delta^T (B + \lambda I) \delta
\end{eqnarray}
Damping the quadratic through $\lambda$ gives a measure of how conservative the quadratic approximation is. A large value of $\lambda$ is more conservative and as $\lambda \rightarrow \infty$ updates become similar to stochastic gradient descent. Alternatively, a small $\lambda$ allows for more substantial parameter updates especially along low curvature directions. Martens dynamically adjusts $\lambda$ at each iteration using a Levenberg-Marquardt style update based on computing the reduction ratio
\begin{eqnarray}
\rho = (f(\theta + \delta) - f(\theta)) / (M_{\theta}(\delta) - M_{\theta}(0))
\end{eqnarray}
If $\rho$ is sufficiently small or negative, $\lambda$ is increased while if $\rho$ is large then $\lambda$ is decreased. The number of CG iterations used to compute $\delta$ has a dramatic effect on $\rho$ which is further discussed in section 4.1. 

To accelerate CG, Martens makes use of the diagonal pre-conditioner
\begin{eqnarray}
P = \bigg[ \text{diag} \bigg( \sum_{j=1}^m \nabla f^{(j)}(\theta) \odot \nabla f^{(j)}(\theta) \bigg) + \lambda I \bigg]^{\xi}
\end{eqnarray}
where $f^{(j)}(\theta)$ is the value of $f$ for datapoint $j$ and $\odot$ denotes component-wise multiplication. $P$ can be easily computed on the same backward pass as computing $\nabla f$. 

Finally, two backtracking methods are used: one after optimizing CG to select $\delta$ and the other a backtracking linesearch to compute the rate $\alpha$. Both these methods operate in the standard way, backtracking through proposals until the objective no longer decreases. 

\section{Stochastic Hessian-free optimization}

Martens' implementation utilizes the full dataset for computing objective values and gradients, and mini-batches for computing curvature-vector products. Naively setting both batch sizes to be small causes several problems. In this section we describe these problems and our contributions in modifying Martens' original algorithm to this setting.

\subsection{Short CG runs, $\delta$-momentum and use of mini-batches}

The CG termination criteria used by Martens is based on a measure of relative progress in optimizing $\hat{M}_{\theta}$. Specifically, if $x_j$ is the solution at CG iteration $j$, then training is terminated when
\begin{eqnarray}
\frac{\hat{M}_{\theta}(x_j) - \hat{M}_{\theta}(x_{j-k})}{\hat{M}_{\theta}(x_j)} < \epsilon
\end{eqnarray}
where $k = $max$(10, j/10)$ and $\epsilon$ is a small positive constant. The effect of this stopping criteria has a dependency on the strength of the damping parameter $\lambda$, among other attributes such as the current parameter settings. For sufficiently large $\lambda$, CG only requires 10-20 iterations when a pre-conditioner is used. As $\lambda$ decreases, more iterations are required to account for pathological curvature that can occur in optimizing $f$ and thus leads to more expensive CG iterations. Such behavior would be undesirable in a stochastic setting where preference would be put towards having equal length CG iterations throughout training. To account for this, we fix the number of CG iterations to be only 3-5 across training for classification and 25-50 for training deep autoencoders. Let $\zeta$ denote this cut-off. Setting a limit on the number of CG iterations is used by \cite{martens2010deep} and \cite{sutskever2011generating} and also has a damping effect, since the objective function and quadratic approximation will tend to diverge as CG iterations increase \cite{martens2012training}. We note that due to the shorter number of CG runs, the iterates from each solution are used during the CG backtracking step.

A contributor to the success of Martens' HF is the use of information sharing across iterations. At iteration $k$, CG is initialized to be the previous solution of CG from iteration $k-1$, with a small decay. For the rest of this work, we denote this as $\delta$-momentum. $\delta$-momentum helps correct proposed update directions when the quadratic approximation varies across iterations, in the same sense that momentum is used to share gradients. This momentum interpretation was first suggested by \cite{martens2012training} in the context of adapting HF to a setting with short CG runs. Unfortunately, the use of $\delta$-momentum becomes challenging when short CG runs are used. Given a non-zero CG initialization, $\hat{M}_\theta$ may be more likely to remain positive after terminating CG and assuming $f(\theta + \delta) - f(\theta) < 0$, means that the reduction ratio will be negative and thus $\lambda$ will be increased to compensate. While this is not necessarily unwanted behavior, having this occur too frequently will push SHF to be too conservative and possibly result in the backtracking linesearch to reject proposed updates. Our solution is to utilize a schedule on the amount of decay used on the CG starting solution. This is motivated by \cite{martens2012training} suggesting more attention on the CG decay in the setting of using short CG runs. Specifically, if $\delta_k^0$ is the initial solution to CG at iteration $k$, then 
\begin{eqnarray}
\delta_k^0 = \gamma_e \delta_{k-1}^{\zeta}, \hspace{3mm} \gamma_e = \text{min}(1.01 \gamma_{e-1},.99)
\end{eqnarray}
where $\gamma_e$ is the decay at epoch $e$,  $\delta_1^0 = 0$ and $\gamma_1 = 0.5$. While in batch training a fixed $\gamma$ is suitable, in a stochastic setting it is unlikely that a global decay parameter is sufficient. Our schedule has an annealing effect in the sense that $\gamma$ values near 1 are feasible late in training even with only 3-5 CG iterations, a property that is otherwise hard to achieve. This allows us to benefit from sharing more information across iterations late in training, similar to that of a typical momentum method.

A remaining question to consider is how to set the sizes of the gradient and curvature mini-batches. \cite{martens2012training} discuss theoretical advantages to utilizing the same mini-batches for computing the gradient and curvature vector products. In our setting, this may lead to some difficulties. Using same-sized batches allows $\lambda \rightarrow 0$ during training \cite{martens2012training}. Unfortunately, this can become incompatible with our short hard-limit on the number of CG iterations, since CG requires more work to optimize $\hat{M}_{\theta}$ when $\lambda$ approaches zero. To account for this, on classification tasks where 3-5 CG iterations are used, we opt to use gradient mini-batches that are 5-10 times larger than curvature mini-batches. For deep autoencoder tasks where more CG iterations are used, we instead set both gradient and curvature batches to be the same size. The behavior of $\lambda$ is dependent on whether or not dropout is used during training.  Figure 1 demonstrates the behavior of $\lambda$ during classification training with and without the use of dropout. With dropout, $\lambda$ no longer converges to 0 but instead plummets, rises and flattens out. In both settings, $\lambda$ does not decrease substantially as to negatively effect the proposed CG solution and consequently the reduction ratio. Thus, the amount of work required by CG remains consistent late in training. The other benefit to using larger gradient batches is to account for the additional computation in computing curvature-vector products which would make training longer if both mini-batches were small and of the same size. In \cite{martens2010deep}, the gradients and objectives are computed using the full training set throughout the algorithm, including during CG backtracking and the backtracking linesearch. We utilize the gradient mini-batch for the current iteration in order to compute all necessary gradient and objectives throughout the algorithm.

\subsection{Levenberg-Marquardt damping}

Martens makes use of the following Levenberg-Marquardt style damping criteria for updating $\lambda$:
\begin{eqnarray}
{\bf if} \rho > \frac{3}{4}, \lambda \leftarrow \frac{2}{3} \lambda \text{ } {\bf else if} \rho < \frac{1}{4}, \lambda \leftarrow \frac{3}{2} \lambda
\end{eqnarray}
which given a suitable initial value will converge to zero as training progresses. We observed that the above damping criteria is too harsh in the stochastic setting in the sense that $\lambda$ will frequently oscillate, which is sensible given the size of the curvature mini-batches. We instead opt for a much softer criterion, for which lambda is updated as
\begin{eqnarray}
{\bf if} \rho > \frac{3}{4}, \lambda \leftarrow \frac{99}{100} \lambda \text{ } {\bf else if} \rho < \frac{1}{4}, \lambda \leftarrow \frac{100}{99} \lambda
\end{eqnarray}
This choice, although somewhat arbitrary, is consistently effective. Thus reduction ratio values computed from curvature mini-batches will have less overall influence on the damping strength. 

\begin{figure}
     \begin{center}

        \subfigure{%
            \includegraphics[width=0.5\textwidth]{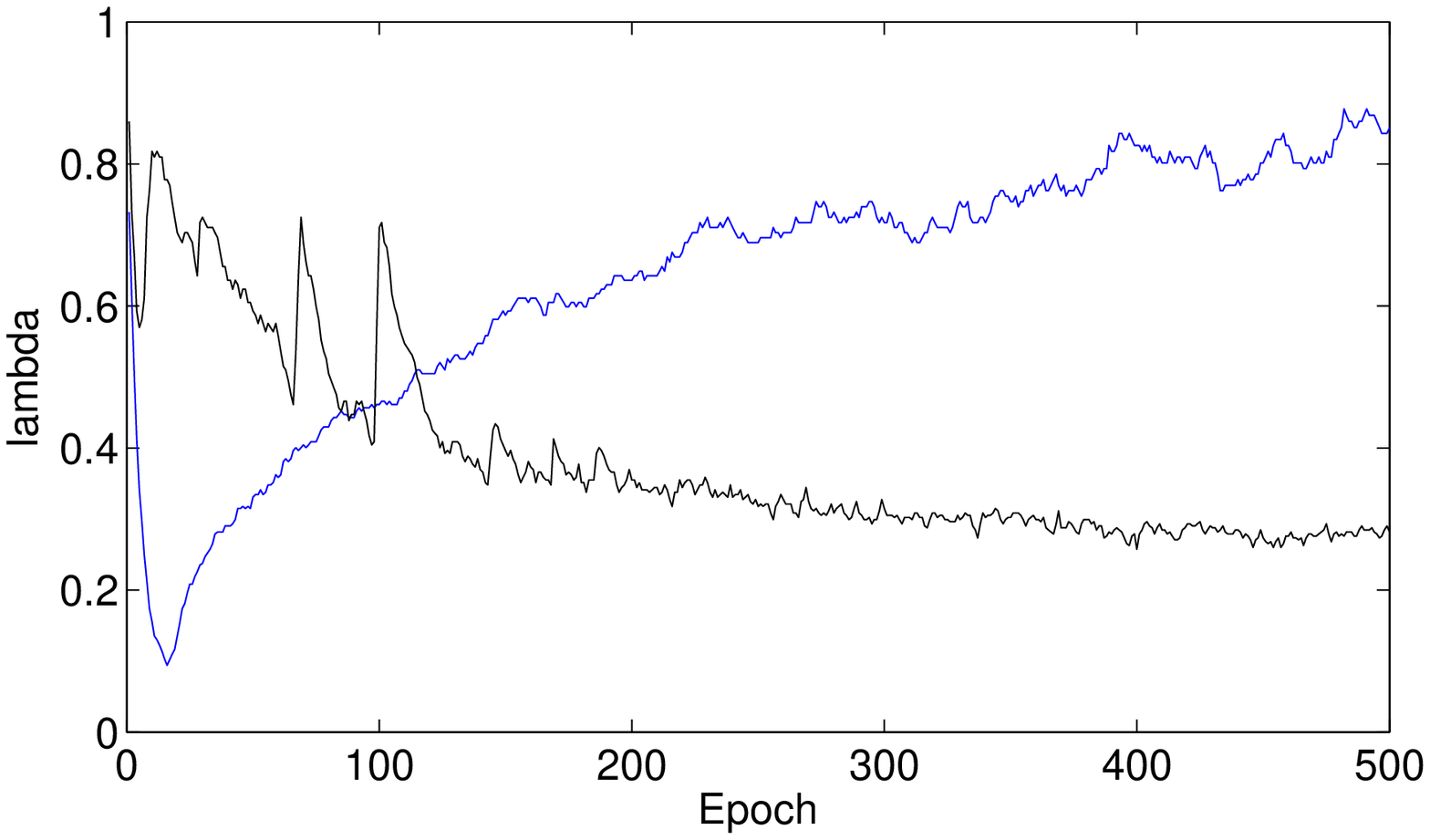}
        }%
        \subfigure{%
            \includegraphics[width=0.5\textwidth]{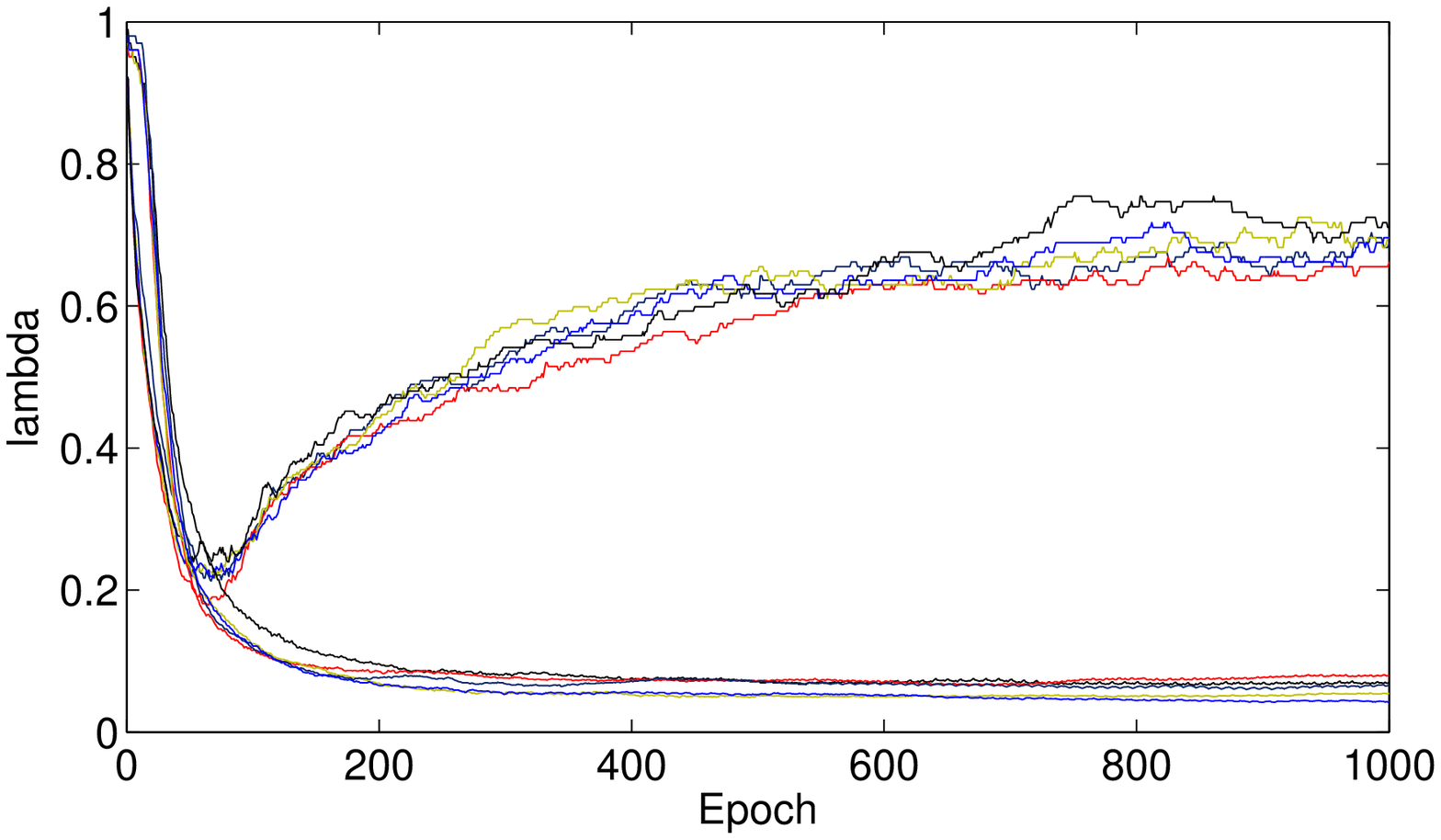}
        }%

    \end{center}
    \caption{%
        Values of the damping strength $\lambda$ during training of MNIST (left) and USPS (right) with and without dropout using $\lambda = 1$ for classification. When dropout is included, the damping strength initially decreases followed by a steady increase over time.
     }%
   \label{fig:extract}
\end{figure}

\subsection{Integrating dropout}

Dropout is a recently proposed method for improving the training of neural networks. During training, each hidden unit is omitted with a probability of 0.5 along with optionally omitting input features similar to that of a denoising autoencoder \cite{vincent2008extracting}. Dropout can be viewed in two ways. By randomly omitting feature detectors, dropout prevents co-adaptation among detectors which can improve generalization accuracy on held-out data. Secondly, dropout can be seen as a type of model averaging. At test time, outgoing weights are halved. If we consider a network with a single hidden layer and $k$ feature detectors, using the mean network at test time corresponds to taking the geometric average of $2^k$ networks with shared weights. Dropout is integrated in stochastic HF by randomly omitting feature detectors on both gradient and curvature mini-batches from the last hidden layer during each iteration. Since we assume that the curvature mini-batches are a subset of the gradient mini-batches, the same feature detectors are omitted in both cases. 

Since the curvature estimates are noisy, it is important to consider the stability of updates when different stochastic networks are used in each computation. The weight updates in dropout SGD are augmented with momentum not only for stability but also to speed up learning. Specifically, at iteration $k$ the parameter update is given by
\begin{eqnarray}
\Delta \theta_k = p_k \Delta \theta_{k-1} - (1 - p_k) \alpha_k \langle \nabla f \rangle, \hspace{3mm} \theta_k = \theta_{k-1} + \Delta \theta_k
\end{eqnarray}
where $p_k$ and $a_k$ are the momentum and learning rate, respectively. We incorporate an additional exponential decay term $\beta_e$ when performing parameter updates. Specifically, each parameter update is computed as
\begin{eqnarray}
\theta_k = \theta_{k-1} + \beta_e \alpha_k \delta_k, \hspace{3mm} \beta_e = c \beta_{e-1}
\end{eqnarray}
where $c \in (0, 1]$ is a fixed parameter chosen by the user. Incorporating $\beta_e$ into the updates, along with the use of $\delta$-momentum, leads to more stable updates and fine convergence particularly when dropout is integrated during training.




\subsection{Algorithm}

Pseudo-code for one iteration of our implementation of stochastic Hessian-free is presented. Given a gradient minibatch $X_k^g$ and curvature minibatch $X_k^c$, we first sample dropout units (if applicable) for the inputs and last hidden layer of the network. These take the form of a binary vector, which are multiplied component-wise by the activations $y_i$. In our pseudo-code, CG$(\delta_k^0, \nabla f, P, \zeta)$ is used to denote applying CG with initial solution $\delta_k^0$, gradient $\nabla f$, pre-conditioner $P$ and $\zeta$ iterations. Note that, when computing $\delta$-momentum, the $\zeta$-th solution in iteration $k-1$ is used as opposed to the solution chosen via backtracking. Given the objectives $f_{k-1}$ computed with $\theta$ and $f_k$ computed with $\theta + \delta_k$, the reduction ratio $\rho$ is calculated utilizing the un-damped quadratic approximation $M_{\theta}(\delta_k)$. This allows updating $\lambda$ using the Levenberg-Marquardt style damping. Finally, a backtracking linesearch with at most $\omega$ steps is performed to compute the rate and serves as a last defense against potentially poor update directions. 

\begin{algorithm}
\caption{Stochastic Hessian-Free Optimization}
\begin{algorithmic}
\STATE $X_k^g \leftarrow$ gradient minibatch, $X_k^c \leftarrow$ curvature minibatch, $|X_k^g| = h|X_k^c|, h \in \mathbb{Z}^+$
\STATE Sample dropout units for inputs and last hidden layer
\IF{start of new epoch}
\STATE $\gamma_e \leftarrow$ min$(1.01 \gamma_{e-1},.99)$ \COMMENT{\{$\delta$-momentum\}}
\ENDIF
\STATE $\delta_k^0 \leftarrow \gamma_e \delta_{k-1}^{\zeta}$
\STATE $f_{k-1} \leftarrow f(X_k^g; \theta)$, $\nabla f \leftarrow \nabla f(X_k^g; \theta),$ $P \leftarrow$ Precon($X_k^g; \theta)$ 
\STATE Solve $(B + \lambda I) \delta_k = -\nabla f$ using CG$(\delta_k^0, \nabla f, P, \zeta)$ \COMMENT{\{Using $X_k^c$ to compute $B \delta_k$\}}
\STATE $f_k \leftarrow f(X_k^g; \theta + \delta_k)$ \COMMENT{\{CG backtracking\}}
\FOR{$j = \zeta$ - 1 to 1}
\STATE $f(\theta + \delta_k^j) \leftarrow f(X_k^g; \theta + \delta_k^j)$
\IF{$f(\theta + \delta_k^j) < f_k$}
\STATE $f_k \leftarrow f(\theta + \delta_k^j), \delta_k \leftarrow \delta_k^j$
\ENDIF
\ENDFOR
\STATE $\rho \leftarrow (f_k - f_{k-1}) / (\frac{1}{2} \delta_k^T B \delta_k + \nabla f^T \delta_k)$ \COMMENT{\{Using $X_k^c$ to compute $B \delta_k$\}}
\STATE {\bf if} $\rho < .25$, $\lambda \leftarrow 1.01 \lambda$  {\bf elseif}  $\rho > .75$, $\lambda \leftarrow .99 \lambda$ {\bf end if}
\STATE $\alpha_k \leftarrow 1, j \leftarrow 0$ \COMMENT{\{Backtracking linesearch\}}
\WHILE{$j < \omega$}
\STATE {\bf if} $f_k > f_{k-1} + .01 \alpha_k \nabla f^T \delta_k$ {\bf then} $\alpha_k \leftarrow .8 \alpha_k, j \leftarrow j + 1$ {\bf else break end if}
\ENDWHILE
\STATE $\theta \leftarrow \theta + \beta_e \alpha_k \delta_k, k \leftarrow k + 1$ \COMMENT{\{Parameter update\}}

\end{algorithmic}
\end{algorithm}

Since curvature mini-batches are sampled from a subset of the gradient mini-batch, it is then sensible to utilize different curvature mini-batches on different epochs. Along with cycling through gradient mini-batches during each epoch, we also cycle through curvature subsets every $h$ epochs, where $h$ is the size of the gradient mini-batches divided by the size of the curvature mini-batches. For example, if the gradient batch size is 1000 and the curvature batch size is 100, then curvature mini-batch sampling completes a full cycle every 1000/100 = 10 epochs. 

Finally, one simple way to speed up training as indicated in \cite{martens2012training}, is to cache the activations when initially computing the objective $f_k$. While each iteration of CG requires computing a curvature-vector product, the network parameters are fixed during CG and is thus wasteful to re-compute the network activations on each iteration. 

\section{Experiments}

We perform experimental evaluation on both classification and deep autoencoder tasks. The goal of classification experiments is to determine the effectiveness of SHF on test error generalization. For autoencoder tasks, we instead focus just on measuring the effectiveness of the optimizer on the training data. The datasets and experiments are summarized as follows:

\begin{itemize}[leftmargin=*]
\item MNIST: Handwritten digits of size $28 \times 28$ with 60K training samples and 10K testing samples. For classification, we train networks of size 784-1200-1200-10 with rectifier activations. For deep autoencoders, the encoder architecture of 784-1000-500-250-30 with a symmetric decoding architecture is used. Logistic activations are used with a binary cross entropy error. For classification experiments, the data is scaled to have zero mean and unit variance.

\item CURVES: Artificial dataset of curves of size $28 \times 28$ with 20K training samples and 10K testing samples. We train a deep autoencoder using an encoding architecture of 784-400-200-100-50-25-6 with symmetric decoding. Similar to MNIST, logistic activations and binary cross entropy error are used. 

\item USPS: Handwritten digits of size $16 \times 16$ with 11K examples. We perform classification using 5 randomly sampled batches of 8K training examples and 3K testing examples as in \cite{min2009deep} Each batch has an equal number of each digit. Classification networks of size 256-500-500-10 are trained with rectifier activations. The data is scaled to have zero mean and unit variance.

\item Reuters: A collection of 8293 text documents from 65 categories. Each document is represented as a 18900-dimensional bag-of-words vector. Word counts $C$ are transformed to log($1 + C)$ as is done by \cite{hinton2012improving}. The publically available train/test split of is used. We train networks of size 18900-65 for classification due to the high dimensionality of the inputs, which reduces to softmax-regression. 

\end{itemize}

For classification experiments, we perform comparison of SHF with and without dropout against dropout SGD \cite{hinton2012improving}. All classification experiments utilize the sparse initialization of Martens \cite{martens2010deep} with initial biases set to 0.1. The sparse initialization in combination with ReLUs make our networks similar to the deep sparse rectifier networks of \cite{glorot2011deep}. All algorithms are trained for 500 epochs on MNIST and 1000 epochs on USPS and Reuters. We use weight decay of $5 \times 10^{-4}$ for SHF and $2 \times 10^{-5}$ for dropout SHF. A held-out validation set was used for determining the amount of input dropout for all algorithms. Both SHF and dropout SHF use initial damping of $\lambda = 1$, gradient batch size of 1000, curvature batch size of 100 and 3 CG iterations per batch.

Dropout SGD training uses an exponential decreasing learning rate schedule initialized at 10, in combination with max-norm weight clipping \cite{hinton2012improving}. This allows SGD to use larger learning rates for greater exploration early in training. A linearly increasing momentum schedule is used with initial momentum of 0.5 and final momentum of 0.99. No weight decay is used. For additional comparison we also train dropout SGD when dropout is only used in the last hidden layer, as is the case with dropout SHF.

For deep autoencoder experiments, we use the same experimental setup as in Chapter 7 of \cite{sutskever2013training}. In particular, we focus solely on training error without any L2 penalty in order to determine the effectiveness of the optimizer on modeling the training data. Comparison is made against SGD, SGD with momentum, HF and Nesterov's accelerated gradient (NAG). On CURVES, SHF uses an initial damping of $\lambda = 10$, gradient and curvature batch sizes of 2000 and 25 CG iterations per batch. On MNIST, we use initial $\lambda = 1$, gradient and curvature batch sizes of 3000 and 50 CG iterations per batch. Autoencoder training is ran until no sufficient progress is made, which occurs at around 250 epochs on CURVES and 100 epochs on MNIST. 

\subsection{Classification results}

\begin{figure}
     \begin{center}

        \subfigure{%
            \includegraphics[width=0.33\textwidth]{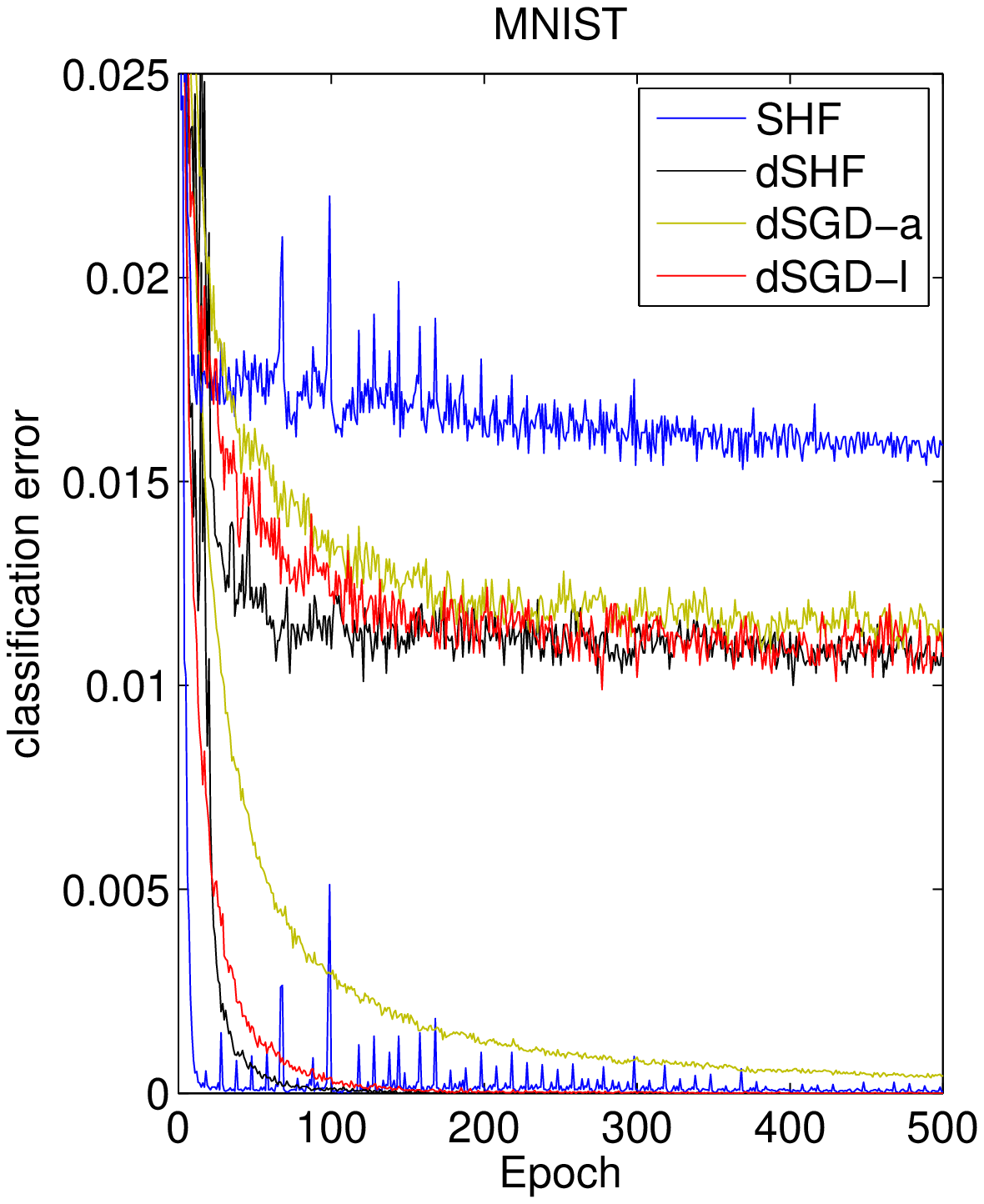}
        }%
        \subfigure{%
            \includegraphics[width=0.332\textwidth]{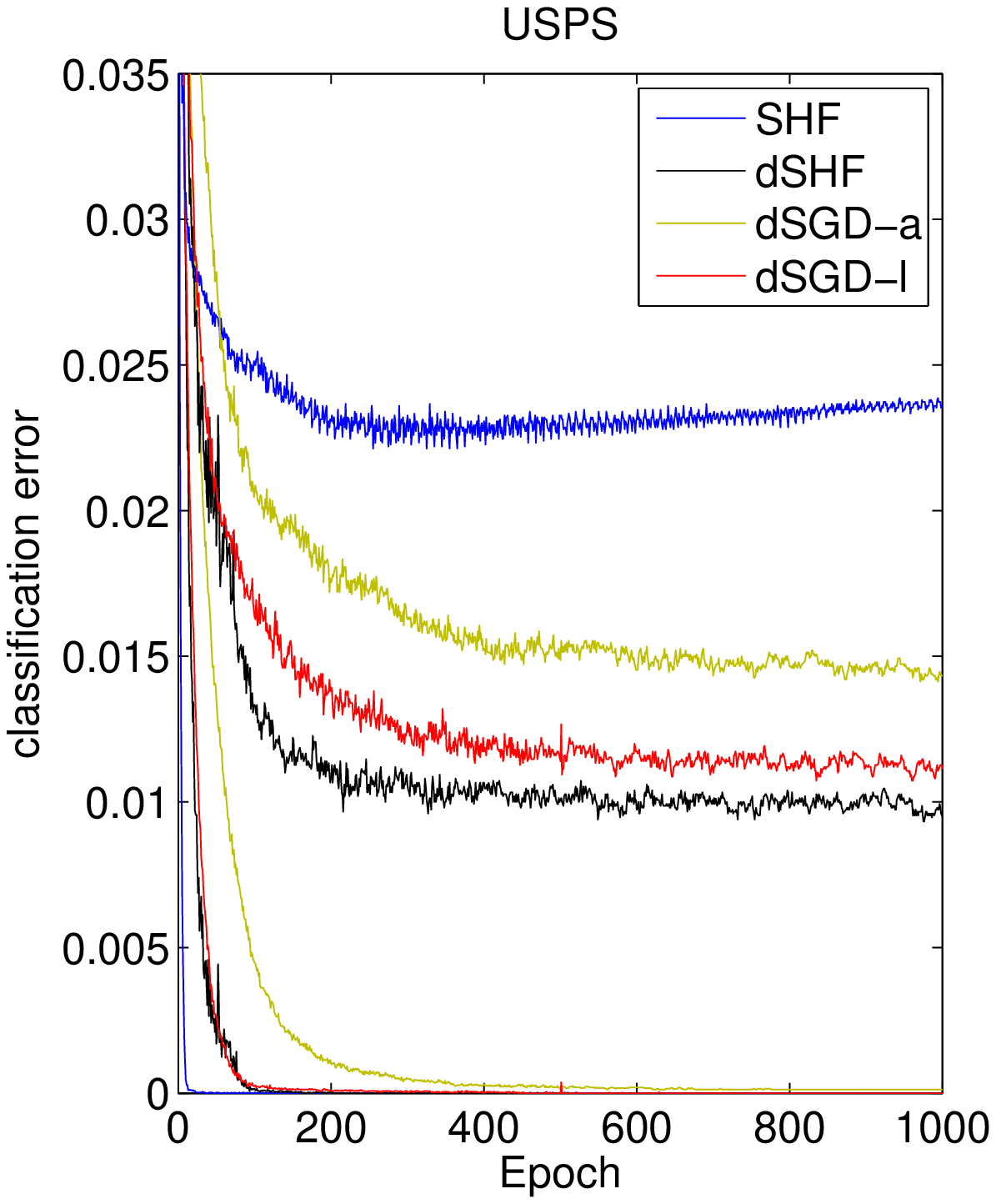}
        }%
        \subfigure{%
           \includegraphics[width=0.326\textwidth]{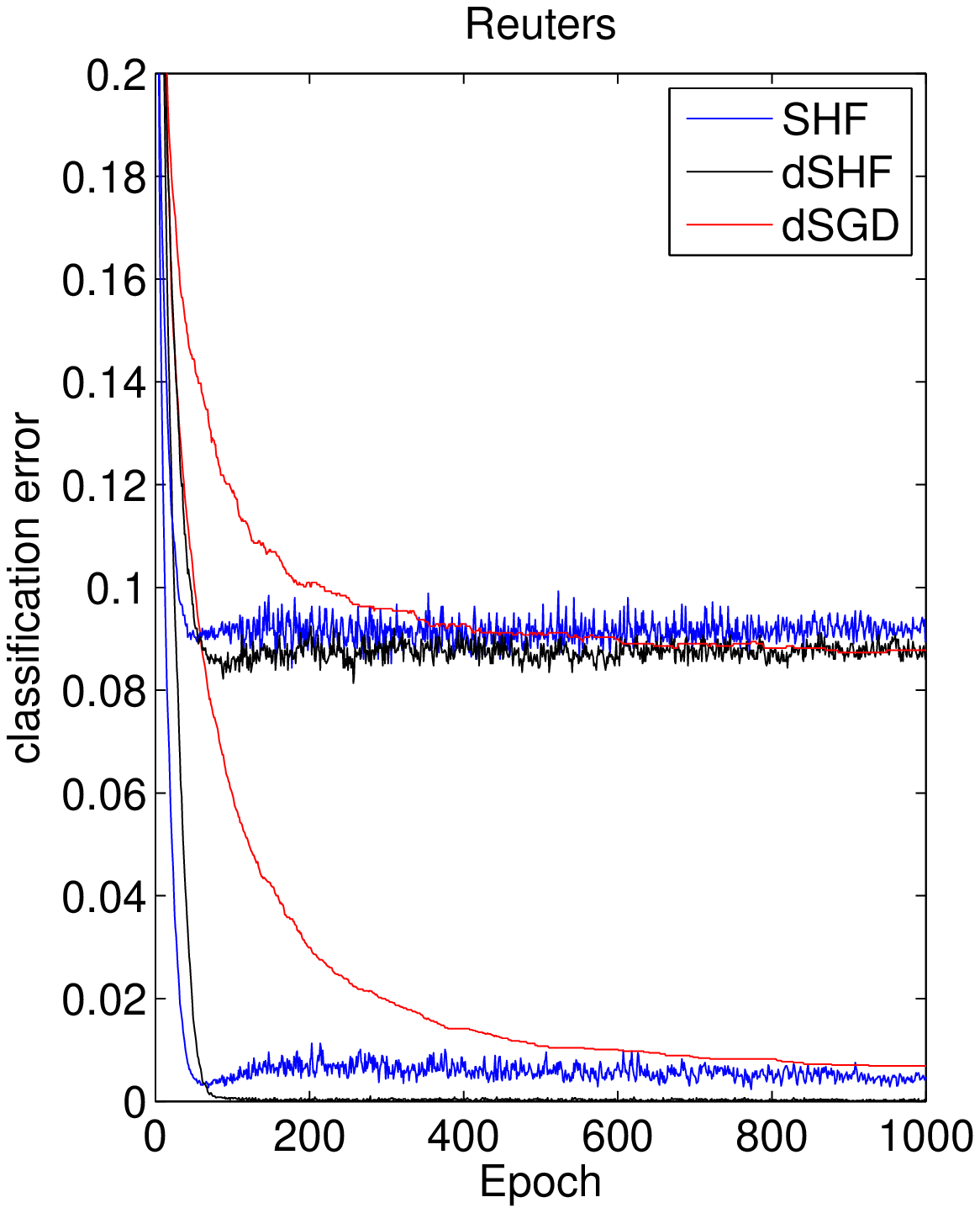}
        }\\ 

    \end{center}
    \caption{%
       Training and testing curves for classification. dSHF: dropout SHF, dSGD: dropout SGD, dSGD-a: dropout on all layers, dSGD-l: dropout on last hidden layer only (as well as the inputs).
     }%
   \label{fig:extract}
\end{figure}

Figure $\ref{fig:extract}$ summarizes our classification results. At epoch 500, dropout SHF achieves 107 errors on MNIST. This result is similar to \cite{hinton2012improving} which achieve 100-115 errors with various network sizes when training for a few thousand epochs. Without dropout or input corruption, SHF achieves 159 errors on MNIST, on par with existing methods that do not incorporate prior knowledge, pre-training, image distortions or dropout. As with \cite{martens2010deep}, we hypothesize that further improvements can be made by fine-tuning with SHF after unsupervised layerwise pre-training. 

After 1000 epochs of training on five random splits of USPS, we obtain final classification errors of 1\%, 1.1\%, 0.8\%, 0.9\% and 0.97\% with a mean test error of 0.95\%. Both algorithms use 50\% input corruption. For additional comparison, \cite{min2009deepp} obtains a mean classification error of 1.14\% using a pre-trained deep network for large-margin nearest neighbor classification with the same size splits. Without dropout, SHF overfits the training data.

On the Reuters dataset, SHF with and without dropout both demonstrate accelerated training. We hypothesize that further speedup may also be obtained by starting training with a much smaller $\lambda$ initialization, which we suspect is conservative given that the problem is convex.

\subsection{Deep autoencoder results}

\begin{table}
\caption{Training errors on the deep autoencoder tasks. All results are obtained from \cite{sutskever2013training}. M(0.99) refers to momentum capped at 0.99 and similarily for M(0.9). SGD-VI refers to SGD using the variance normalized initialization of \cite{glorot2010understanding}. }
\begin{center}
\begin{tabular}{ l | c c c c c c | c}
problem & NAG & M(0.99) & M(0.9) & SGD & SGD-VI \cite{chapelle2011improved} & HF & SHF \\
\hline 
CURVES & 0.078 & 0.110 & 0.220 & 0.250 & 0.160 & 0.110 & 0.089 \\
MNIST & 0.730 & 0.770 & 0.990 & 1.100 & 0.900 & 0.780 & 0.877 \\
\end{tabular}
\end{center}
\label{tab:auto}
\end{table}

\begin{figure}
     \begin{center}

        \subfigure{%
            \includegraphics[width=0.359\textwidth]{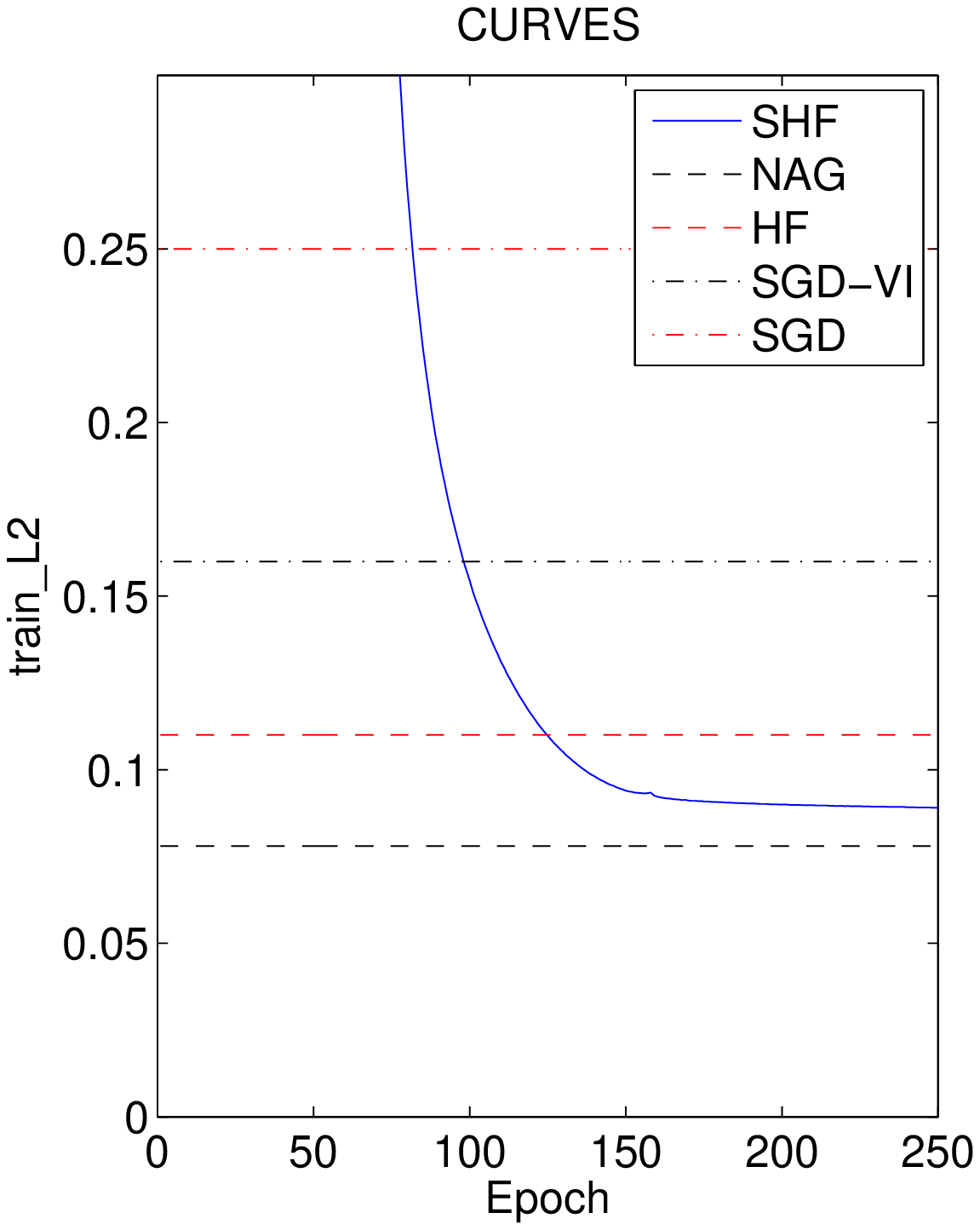}
        }%
        \subfigure{%
            \includegraphics[width=0.35\textwidth]{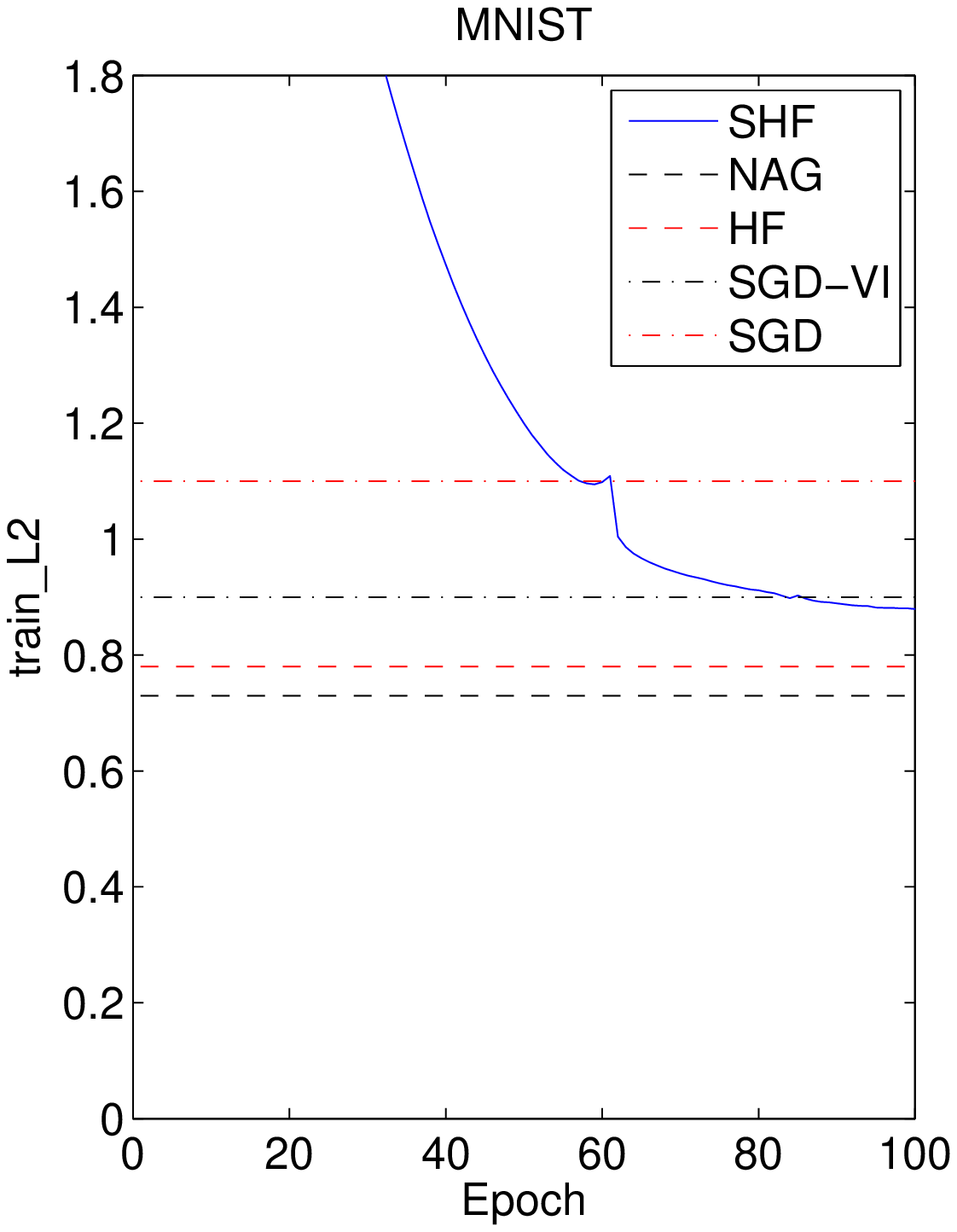}
        }%

    \end{center}
    \caption{%
        Learning curves for the deep autoencoder tasks. The CG decay parameter $\gamma$ is shut off at epoch 160 on CURVES and epoch 60 on MNIST.
     }%
   \label{fig:auto}
\end{figure}

Figure $\ref{fig:auto}$ and table $\ref{tab:auto}$ summarize our results. Inspired by \cite{sutskever2013training} we make one additional modification to our algorithms. As soon as training begins to diverge, we turn off the CG decay parameter $\gamma$ in a similar fashion as the the momentum parameter $\mu$ is decreased in \cite{sutskever2013training}. When $\gamma = 0$, CG is no longer initialized from the previous solution and is instead initialized to zero. As with \cite{sutskever2013training}, this has a dramatic effect on the training error but to a lesser extent as momentum and Nesterov's accelerated gradient. \cite{sutskever2013training} describes the behaviour of this effect as follows: with a large momentum, the optimizer is able to make steady progress along slow changing directions of low curvature. By decreasing the momentum late in training, the optimizer is then able to quickly reach a local minimum from finer optimization along high curvature directions, which would otherwise be too difficult to obtain with an aggressive momentum schedule. This observation further motivates the relationship between momentum and information sharing through CG.

Our experimental results demonstrate that SHF does not perform significantly better or worse on these datasets compared to existing approaches. It is able to outperform HF on CURVES but not on MNIST. An attractive property that is shared with both HF and SHF is not requiring the careful schedule tuning that is necessary for momentum and NAG. We also attempted experiments with SHF using the same setup for classification with smaller batches and 5 CG iterations. The results were worse: on CURVES the lowest training error obtained was 0.19. This shows that while such a setup is useful from the viewpoint of noisy updates and test generalization, they hamper the effectiveness of making progress on hard to optimize regions.

\section{Conclusion}

In this paper we proposed a stochastic variation of Martens' Hessian-free optimization incorporating dropout for training neural networks on classification and deep autoencoder tasks. By adapting the batch sizes and number of CG iterations, SHF can be constructed to perform well for classification against dropout SGD or optimizing deep autoencoders comparing HF, NAG and momentum methods. While our initial results are promising, of interest would be adapting stochastic Hessian-free optimization to other network architectures:

\begin{itemize}[leftmargin=*]
\item {\bf Convolutional networks.} The most common approach to training convolutional networks has been SGD incorporating a diagonal Hessian approximation \cite{lecun1998efficient}. Dropout SGD was recently used for training a deep convolutional network on ImageNet \cite{krizhevsky2012imagenet}.
\item {\bf Recurrent Networks.} It was largely believed that RNNs were too difficult to train with SGD due to the exploding/vanishing gradient problem. In recent years, recurrent networks have become popular again due to several advancements made in their training \cite{bengio2012advances}.
\item {\bf Recursive Networks.} Recursive networks have been successfully used for tasks such as sentiment classification and compositional modeling of natural language from word embeddings \cite{socher2012semantic}. These architectures are usually trained using L-BFGS.  
\end{itemize}

It is not clear yet whether this setup is easily generalizable to the above architectures or whether improvements need to be considered. Furthermore, additional experimental comparison would involve dropout SGD with the adaptive methods of Adagrad \cite{duchi2010adaptive} or \cite{schaul2012no}, as well as the importance of pre-conditioning CG. None the less, we hope that this work initiates future research in developing stochastic Hessian-free algorithms. 

\subsubsection*{Acknowledgments}

The author would like to thank Csaba Szepesv\'{a}ri for helpful discussion as well as David Sussillo for his guidance when first learning about and implementing HF. The author would also like to thank the anonymous ICLR reviewers for their comments and suggestions.

{\bibliography{iclr2013}}
{\bibliographystyle{unsrt}}

\end{document}